\def\paperTitle{How Much Temporal Long-Term Context is Needed for Action Segmentation?}

\def\authorBlock{
    Emad Bahrami$^1$  \qquad
    Gianpiero Francesca$^2$ \qquad
    Juergen Gall$^{1,3}$ \\
    \normalsize{$^1$University of Bonn, Germany} \qquad
    \normalsize{$^2$Toyota Motor Europe, Belgium} \\ 
    \normalsize{$^3$Lamarr Institute for Machine Learning and Artificial Intelligence, Germany}
}

\newif\ifreview 
\newif\ifarxiv 
\newif\ifcamera \newcommand{\cameraready}{\cameratrue}
\newif\ifrebuttal

\pdfoutput=1
\documentclass[10pt,twocolumn,letterpaper]{article}

\usepackage{iccv}

\usepackage{times}
\usepackage{microtype}
\usepackage{epsfig}
\usepackage[table,xcdraw]{xcolor}
\usepackage{caption}
\usepackage{float}
\usepackage{placeins}
\usepackage{stfloats}
\usepackage{enumitem}
\usepackage{tabularx}
\usepackage{xstring}
\usepackage{multirow}
\usepackage{xspace}

\usepackage{subcaption}
\usepackage{graphicx}
\usepackage{amsmath}
\usepackage{amssymb}
\usepackage{booktabs}
\usepackage{pifont}
\usepackage[hang,flushmargin]{footmisc}
\usepackage{url}

\def\eg{\emph{e.g}.} 
\def\ie{\emph{i.e}.}

\def\etal{\emph{et al}.}

\ifarxiv  \fi

\newcommand{\R}[1]{{%
    \textbf{%
        \ifstrequal{#1}{1}{\textcolor{red}{R#1}}{%
        \ifstrequal{#1}{2}{\textcolor{blue}{R#1}}{%
        \ifstrequal{#1}{3}{\textcolor{magenta}{R#1}}{%
        \ifstrequal{#1}{4}{\textcolor{teal}{R#1}}{%
                           \textcolor{cyan}{R#1}%
        }}}}%
    }%
}}

\usepackage[breaklinks=true,bookmarks=false]{hyperref}

\cameraready
\iccvfinalcopy 

\ificcvfinal\fi

\DeclareMathOperator{\softmax}{softmax}

\begin{document}
\title{\paperTitle}
\author{\authorBlock}
\maketitle

\begin{abstract}
Modeling long-term context in videos is crucial for many fine-grained tasks including temporal action segmentation. An interesting question that is still open is how much long-term temporal context is needed for optimal performance. While transformers can model the long-term context of a video, this becomes computationally prohibitive for long videos. Recent works on temporal action segmentation thus combine temporal convolutional networks with self-attentions that are computed only for a local temporal window. While these approaches show good results, their performance is limited by their inability to capture the full context of a video. In this work, we try to answer how much long-term temporal context is required for temporal action segmentation by introducing a transformer-based model that leverages sparse attention to capture the full context of a video. We compare our model with the current state of the art on three datasets for temporal action segmentation, namely 50Salads, Breakfast, and Assembly101. Our experiments show that modeling the full context of a video is necessary to obtain the best performance for temporal action segmentation.
 
\end{abstract}
\section{Introduction}
\label{sec:intro}
\begin{figure}[t]
    \centering
    \includegraphics[width=1.0\columnwidth]{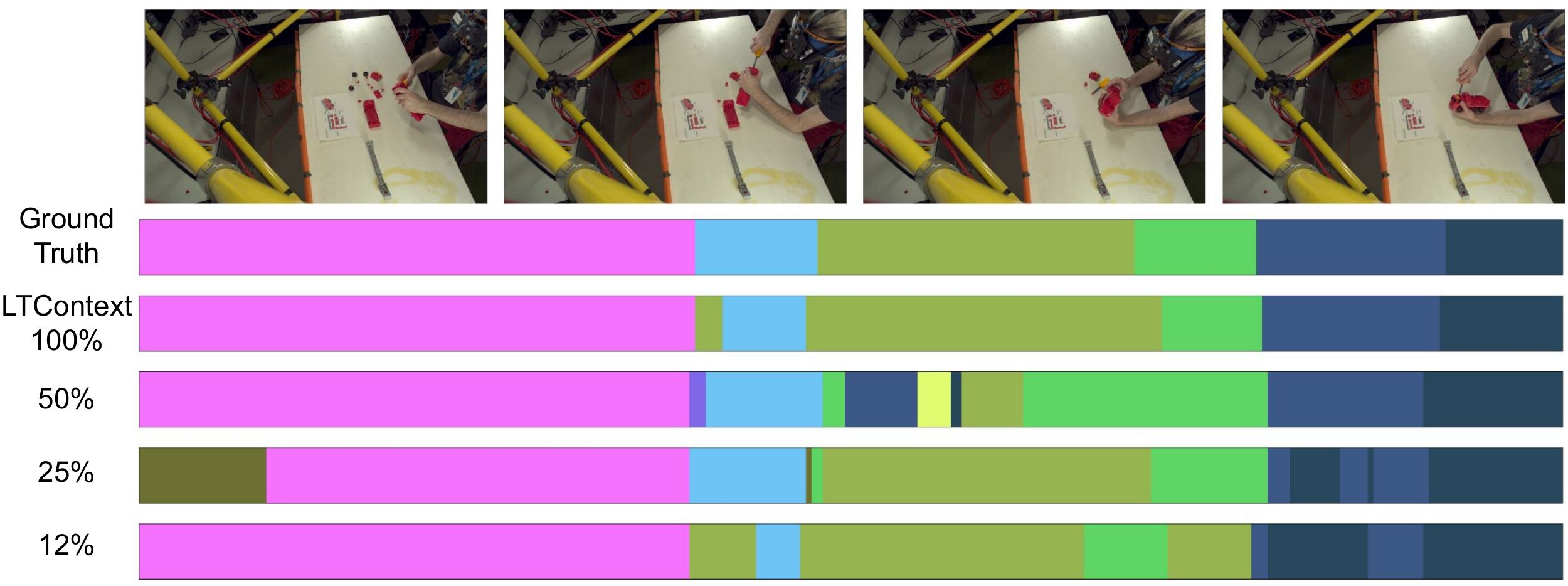}
    \caption{Datasets like Assembly101 contain long videos of assembly tasks and an action label needs to be predicted for each frame. The first row shows some frames of a video. The second row shows the ground-truth labels for all frames of the video where different colors correspond to different action labels. Rows 3-6 show the predictions of the proposed model for different amounts of long-term context where 100\% means the temporal context of the full video.          
    }
    \label{fig:teaser_assembly}
\end{figure}

Temporal action segmentation can be used in many real-world applications such as monitoring production lines or studying animal behavior. In these settings, the videos can be very long and it is required to recognize the start and end of all actions that occur in a video as illustrated in Fig.~\ref{fig:teaser_assembly}.

Recently, combinations of temporal convolutional networks~\cite{farha2019ms,li2020ms} with self- and cross-attention from transformers~\cite{yi2021asformer, behrmann2022uvast} have shown impressive results for temporal action segmentation. These works are in line with other hybrid models~\cite{gulati2020, dai2021coatnet, yi2021asformer, xiao2021} that combine the attention modules with convolutions to compensate for the lack of strong inductive bias of pure transformers. However, the emergence of datasets like Assembly101~\cite{sener2022assembly101}, where subjects perform assembly tasks, poses a new challenge in the area of temporal action segmentation due to the existence of long videos that can last up to 25 minutes. Since modeling the long-term context of such a video is very expensive, Yi \etal~\cite{yi2021asformer} proposed to compute the attention for a local temporal window. In order to understand the impact of the window size on the temporal action segmentation accuracy, we analyze the impact of the window size on two datasets with long videos in Section \ref{sec:eval}. Fig.~\ref{fig:teaser_assembly} shows some qualitative results of this study on Assembly101. Indeed, the results show that modeling the long-term context of an entire video is very important for temporal action segmentation. 

Based on this finding, we revisit how temporal attention is modeled in transformer architectures for temporal action segmentation. \cite{yi2021asformer, behrmann2022uvast} use a hierarchy of temporal windows, making training on long video sequences as they occur in Assembly101~\cite{sener2022assembly101} very expensive. Inspired by works that decompose attention over the spatial and temporal domain for short video clips~\cite{bertasius2021TimeSformer, zhang2021vidtr}, we propose to iterate between computing windowed local attention and sparse long-term context attention such that both short and long-term context is modeled. This approach is particularly suitable for temporal action segmentation since the local attentions focus on the similarity or dissimilarity of features within an action segment or between neighboring action segments whereas the long-term context attention focuses on the relations between actions within the entire video. The source code is available at \url{https://github.com/LTContext/LTContext}.

\section{Related Work}
\label{sec:related}

The traditional sliding window with non-maxima suppression has been among the early approaches for action segmentation~\cite{karaman2014, rohrbach2012}. \cite{kuehne2016end, kuehne2014language, tang2012latent} adopted hidden Markov Models (HMMs) for temporal modeling. \cite{richard2016temporal} incorporates length and temporal context and uses dynamic programming for inference. Temporal convolutional networks (TCN) with temporal pooling are used by~\cite{lea2017temporal} to classify each video frame. Later, \cite{farha2019ms} introduced a multi-stage TCN capable of maintaining high temporal resolution, which is necessary for fine-grained recognition. \cite{ishikawa2021asrf} reduces the over-segmentation error by adding an action boundary regression network to refine the frame-wise results. \cite{huang2020improving} proposed another refinement method based on a graph convolutional network. 

Transformers~\cite{vaswani2017attention} were originally used for natural language processing and only rely on attention for sequence modeling. Recently, transformers have been widely adopted in vision~\cite{dosovitskiy2021vit, wu2021cvt, liu2021Swin, wang2021panoptic}, speech recognition~\cite{gulati2020, peng2022branch}, and action recognition~\cite{bertasius2021TimeSformer, arnab2021vivit}. The original transformers suffered from two issues. First, the cost of the self-attention operation is quadratic with respect to the sequence length. This has been addressed by many methods that improve the memory efficiency of transformers~\cite{beltagy2020longformer, nawrot2021hierarchical, dai2019transformer}. We refer to \cite{tay2022survey} for a comprehensive survey of efficient transformers. For instance, restricting the attention to a fixed window size~\cite{beltagy2020longformer} is one approach. Multi-axis self-attention~\cite{tu2022maxvit} combines block-attention with grid-attention, which is based on a spatial grid overlaid on the entire 2D space. Adapting sparse attention for capturing global information has shown to be an effective solution in vision tasks such as high-resolution image generation~\cite{zhao2021improved, tu2022maxim}, object detection~\cite{tu2022maxvit, yang2021focalattn}, and instance segmentation~\cite{tu2022maxvit, yang2021focalattn}. MViT~\cite{fan2021mvit} uses a pooling operation to reduce the space-time resolution before attention and MViTv2~\cite{li2022mvitv2} improves MViT by adding the relative position along (x,y,t) and residual connections. The second issue of transformers is the poor generalization due to a relatively weak inductive bias~\cite{dai2021coatnet, touvron2021training} compared to convolutional neural networks (CNN). Hybrid models that combine self-attention and convolution layers have thus been proposed for vision tasks~\cite{xiao2021, dai2021coatnet, tu2022maxvit, wu2021cvt} and speech recognition~\cite{gulati2020, peng2022branch}. 

In action recognition, works such as \cite{bertasius2021TimeSformer, zhang2021vidtr} adopt the idea of using attention for video understanding. 
In TimesFormer~\cite{bertasius2021TimeSformer}, the self-attention is first applied along the temporal dimension for each single patch (time attention), i.e., over all frames of a short video clip of a few seconds. In a second step,  the attention is applied over the patches of each frame (space attention). VIDTR~\cite{zhang2021vidtr} also decomposes self-attention into spatial and temporal attention, but additionally down-samples the temporal dimension. In our work, we do not apply attention spatially and temporally, but we address the question of how temporal attention can be computed for very long sequences that last 25 minutes as it is required for temporal action segmentation. 

For temporal action segmentation, \cite{yi2021asformer} proposed an architecture, called ASFormer, which is based on a multi-stage TCN~\cite{farha2019ms} and equips the temporal convolutions with a local window attention~\cite{beltagy2020longformer}. This is done in a hierarchy where the size of the local window grows with each layer. 

Recently, \cite{behrmann2022uvast} proposed a decoder on top of the encoder that generates action segments in an autoregressive manner. It uses two heads where the head on top of the encoder predicts frame-wise probabilities and the head of the decoder predicts the sequence of actions. Finally, an alignment decoder fuses the output of the two heads and aligns the predicted sequence of actions to the frames. \cite{aziere2022multistage} introduced a hybrid Temporal Convolution Transformer (TCTr) where they adapt an action boundary detector to adaptively estimate attention from local neighboring frames. \cite{xudon2022dtl} proposed to use additional constraints during training, but the approach assumes that action sequences can be modelled by a directed acyclic graph, which does not allow that actions occur more than once in a video. Recently, multi-modal approaches that combine language models with vision transformers have been proposed. For instance, \cite{li2022bridge} uses prompt engineering to extract features from pre-trained vision-language models such as ActionCLIP~\cite{wang2021actionclip}. 

\section{Long-Term Context for Action Segmentation}
\label{sec:method}

Recently, transformers combined with temporal convolutional neural networks~\cite{yi2021asformer,behrmann2022uvast,aziere2022multistage} have shown very good performance for temporal action segmentation. For this task, the frame-wise labels $c_{1}, \ldots, c_{T}$, where $c_t \in \mathcal{C}$ and $C{=}\vert\mathcal{C}\vert$ denotes the number of action classes, need to be predicted for a given a video $X = (x_{1}, \ldots, x_{T} )$ with $T$ frames, where $x_{t}$ represents a feature map of size $D$ at frame $t$. Since $T$ can be very large, \cite{yi2021asformer, behrmann2022uvast, aziere2022multistage} limit the self-attention to a local temporal window.   

In order to understand how much temporal long-term context is needed for temporal action segmentation, we limited the temporal input window and evaluated the quality of the temporal action segmentation in Section~\ref{sec:eval}. The results in Fig.~\ref{fig:windowing_expr} show that temporal long-term context has a strong impact on the performance. Based on our analysis, we thus revisit the windowed attention of previous works for temporal action segmentation and propose to model the temporal long-term context of a video using sparse attentions~\cite{beltagy2020longformer, qiu2020blockwise, parmar2018imgtransformer}. Additionally, we equip our method with windowed attention to capture the locality between neighboring frames.
In this way, we obtain a flexible design that is capable of providing long-term and local temporal context. 
While we describe first the Long-Term Context (LTContext) block in Section~\ref{sec:global_local_attention}, the entire network is described in Section~\ref{sec:deglas_arch}.

\begin{figure*}[t]
    \centering
    \includegraphics[width=1.0\linewidth]{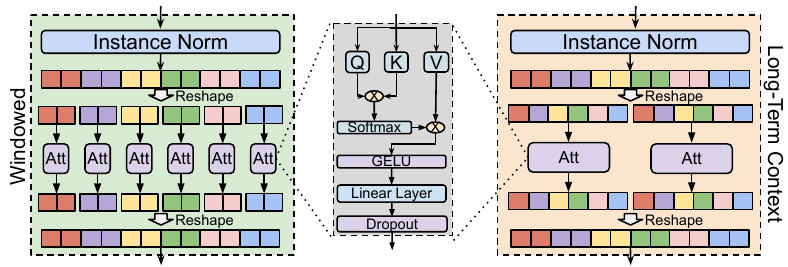}
    \caption{Illustration of windowed and long-term context attentions with a window of size 2. For the windowed attention, the sequence is partitioned into small windows and the attentions are computed for each window. For long-term context attention, the sequence is reordered such that the attentions are computed over the whole, but sparsely sampled sequence. After the attention, the output is reordered again to preserve the original order. Best viewed in color.}
    \label{fig:globa_local_attn}
\end{figure*}

\subsection{Temporal Context Attention}
\label{sec:global_local_attention}
The self-attention blocks of transformers are advantageous over convolutions in the aggregation of global information. However, applying attention to long sequences such as untrimmed videos is impractical due to the quadratic complexity of the self-attention blocks. To address this issue, we adopt an attention mechanism where we leverage sparse and windowed attention to model long-term and local temporal context.

The attention function transforms the input into a query, key, and value and computes the output as a weighted sum of the values. For example, given a sequence of features, $X\in \mathbb{R}^{T \times D}$, the attention can be written as follows:   
\begin{equation}
   Attention(Q, K, V) = \softmax_{row}\left(\frac{QK^{T}}{\sqrt{D}}\right)V 
   \label{eq:attention}
\end{equation}
 where $Q, K, V\in\mathbb{R}^{T \times D}$ are linearly transformed from $X$. Since $T$ is very large for long sequences, we need to modify $Q, K, V$ to enable the modeling of long-term and local temporal context as illustrated in Fig.~\ref{fig:globa_local_attn}.

\paragraph{Temporal Windowed Attention} For the temporal Windowed Attention (WA), we partition the sequence into non-overlapping windows of size $W$. Fig.~\ref{fig:globa_local_attn} illustrates the case for $W=2$, but we use $W=64$ in practice. The impact of $W$ is evaluated in Section \ref{sec:expr}. Instead of computing the attention over the entire sequence of length $T$, we compute the attention $\frac{T}{W}$ times, where each query $Q\in\mathbb{R}^{W \times D}$ corresponds to each window. For the keys and values, we use an overlap where we use the next window in addition, \ie, for each query $Q$ we have $K, V\in\mathbb{R}^{2W \times D}$. We perform masking when $K$ and $V$ exceed the input sequence. We evaluate the impact of the overlap in Section~\ref{sec:expr}.

 \paragraph{Temporal Long-Term Context Attention} For the temporal Long-Term Context (LTC) attention, the input is also partitioned into non-overlapping windows of size $G$. However, instead of computing the attention over each window, the attention is computed over all windows where from each window only one element is taken. In the illustration of Fig.~\ref{fig:globa_local_attn} with $G=2$, we compute the attention over the first feature of all windows and the attention over the second feature of all windows. In general, we compute the attentions for ${G}$ queries $Q\in\mathbb{R}^{\frac{T}{G} \times D}$ where the keys and values are the same, \ie, $K, V\in\mathbb{R}^{\frac{T}{G} \times D}$. The parameter $G$ provides the flexibility to adjust the sparseness based on the available memory budget, \eg, $G=1$ corresponds to the case where the attention is applied over the full sequence. In practice, we use $G=64$ and we evaluate the impact of $G$ in Section~\ref{sec:expr}.

\begin{figure*}[t]
    \centering
    \includegraphics[width=1.0\linewidth]{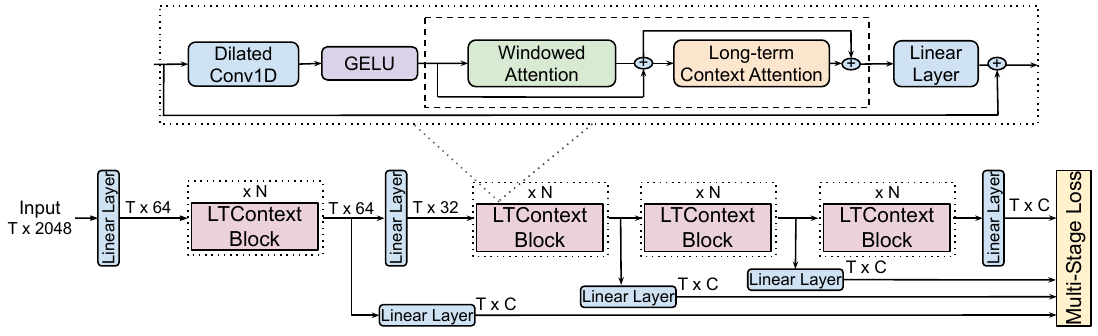}
    \caption{The network architecture of LTContext with LTContext blocks (top).}
    \label{fig:ltc_arch}
\end{figure*}

\paragraph{LTContext block}
The top of Fig.~\ref{fig:ltc_arch} illustrates the entire LTContext block. As in previous works \cite{farha2019ms,yi2021asformer,behrmann2022uvast}, we use a 1D dilated temporal convolution with kernel size 3, where the dilation factor increases by factor $2$ for each layer. The dilated temporal convolution is followed by a Gaussian Error Linear Unit (GELU). In the LTContext block, we first use the windowed and then the long-term context attention, which are shown in Fig.~\ref{fig:globa_local_attn}. We evaluate the impact of the order in Section~\ref{sec:expr}.
Finally, we use a linear layer with a residual connection to output the features for each frame, $F \in \mathbb{R}^{T \times D}$.

\subsection{LTContext Architecture}
\label{sec:deglas_arch}

The entire LTContext network is depicted in Fig.~\ref{fig:ltc_arch}. For a fair comparison, we will use the features that are provided for the corresponding datasets as input. In all cases, the dimensionality of the features is 2048. As in previous works, we use a linear layer to reduce the dimensionality of the features to 64. The output of each LTContext block is the feature map $F \in \mathbb{R}^{T \times D}$. We repeat each LTContext block $N$ times where the dilation factor of the temporal convolution increases in each layer. In practice, we use $N=9$ and we evaluate the impact of $N$ in Section~\ref{sec:expr}. After the first $N$ layers of LTContext blocks, we use an additional linear layer to reduce the dimensionality $D$ further to 32. The dimensionality reduction reduces the number of parameters from $1.42$ million to $0.72$ million without reducing the accuracy. We also use an additional linear layer followed by a softmax layer to generate the frame-wise class probabilities $P \in \mathbb{R}^{T \times C}$. 

We continue with three additional stages where each stage consists of $N$ layers of LTContext blocks. Note that we reset the dilation factor to 1 for the temporal convolution at the beginning of each stage and we compute the frame-wise class probabilities $P \in \mathbb{R}^{T \times C}$ after each stage, which contributes to the multi-stage loss. We use the cross-entropy loss combined with the mean squared error smoothing loss as introduced by~\cite{farha2019ms} and used in \cite{li2020ms,yi2021asformer} for a fair comparison. Inspired by~\cite{yi2021asformer}, we use the cross-attention for the LTContext blocks in stages 2 to 4. Instead of using the features $F$ for the queries and keys for windowed and long-term context attention, the predictions $P$ are used. We thus have $Q\in\mathbb{R}^{W \times C}$, $K\in\mathbb{R}^{2W \times C}$, and $V\in\mathbb{R}^{2W \times D}$ for the windowed attention and $Q\in\mathbb{R}^{\frac{T}{G} \times C}$, $K\in\mathbb{R}^{\frac{T}{G} \times C}$, and $V\in\mathbb{R}^{\frac{T}{G} \times D}$ for the long-term context attention. While the cross-attention is not shown in Fig.~\ref{fig:ltc_arch}, it only means that $P$ is an additional input for the windowed and long-term context attention in stages 2-4. We evaluate the impact of the number of stages in Section~\ref{sec:expr}. 
\section{Experiments}
\label{sec:expr}

\textbf{Datasets.} We evaluate the performance of our proposed model on three challenging action segmentation datasets: 50Salads~\cite{50salads}, Breakfast~\cite{breakfast}, and Assembly101~\cite{sener2022assembly101}.  

\textbf{50Salads}~\cite{50salads} contains 50 videos annotated with 17 action classes. On average, each video is 6.4 minutes long and has 18 action segments. Following previous works~\cite{farha2019ms, yi2021asformer, behrmann2022uvast}, we use five-fold cross-validation and report the average.  

\textbf{Breakfast} contains 1,712 videos of breakfast preparation activities with an average length of 2.3 minutes. There are 48 action classes and each video has on average 6.6 action segments. For evaluation, we report the average of the 4 splits for cross-validation as in~\cite{farha2019ms}.

\textbf{Assembly101}~\cite{sener2022assembly101} is the largest dataset among the three datasets with 4,321 videos and 202 coarse action classes composed of 11 verbs and 61 objects.
Assembly101 is a procedural activity dataset containing videos of people assembling and disassembling 101 toy vehicles. On average, each video includes 24 action segments and is 7.1 minutes long. Compared to Breakfast, Assembly101 has 2.5 times more videos, 6.7 times more hours of video footage, 9.3 times more action segments, and 4.2 more action classes. For our evaluation, we follow the setting for temporal action segmentation~\cite{sener2022assembly101} and report the results on the validation set.   

For a fair comparison, we use the features that are provided for the datasets and that have been used in previous works. We use the I3D~\cite{carreira2017i3d} features for the 50Salads and Breakfast datasets and TSM~\cite{lin2019tsm} features for the Assembly101 dataset~\cite{sener2022assembly101}. Both features are 2048 dimensional vectors. Following~\cite{farha2019ms}, we also used the temporally downsampled features for 50Salads to compensate for the different frame-rates of the datasets. 

\textbf{Evaluation Metrics.} We report the frame-wise accuracy (Acc), segmental Edit distance, and segmental F1 score at the overlapping thresholds of $10\%$, $25\%$, and $50\%$ denoted by $F1 @ \{10, 25, 50 \}$. The intersection over the union (IoU) ratio is used as the overlapping threshold. The edit distance measures only the order of the actions but not the duration. The frame-wise accuracy measures the accuracy per frame. It is dominated by actions that have a long duration and it is not very sensitive to over-segmentation errors that occur when a few frames within a ground-truth action segment are wrongly classified. The F1 score is the most reliable measure.           

\subsection{How much temporal long-term context is needed?}\label{sec:eval}
We first present the results of our analysis on the impact of using the full sequence as input compared to using a temporal window. The goal of this experiment is to shed light on how much temporal long-term context is needed for the task of temporal action segmentation. For the analysis, we use only 50Salads and Assembly101 since the videos in Breakfast are too short. For the experiments, we train our approach (LTContext) and ASFormer \cite{yi2021asformer} either on the full video sequences or we divide the videos into shorter sequences where the full context is lacking. 

\begin{figure}[t]
    \centering
    \includegraphics[width=\columnwidth]{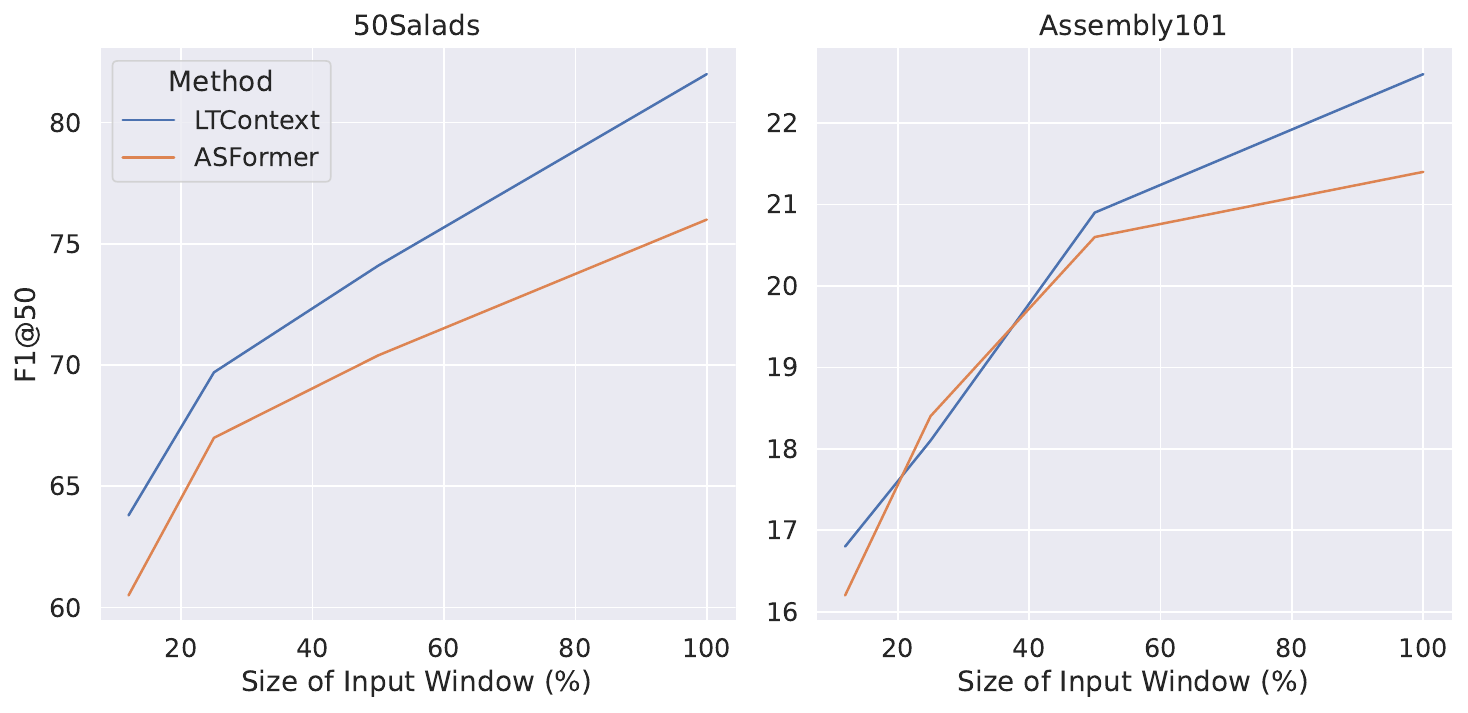}
    \caption{Impact of different sizes of the input window on the 50Salads dataset (left) and the Assembly101 dataset (right). The window size is given in percentage of the average length of a video in the corresponding dataset. 100\% denotes the entire video.}
    \label{fig:windowing_expr}
\end{figure}

Fig.~\ref{fig:windowing_expr} shows the result of this experiment on the 50Salads and Assembly101 dataset. We report the window size as a percentage of the average length of a video in the corresponding dataset and 100\% means that the full sequence has been used. The results clearly show that the full context of an entire sequence is advantageous over allowing the model to see only a window of the input sequence even if the window is large (50\%). We can also observe that our approach benefits more from the full sequence than ASFormer.  

\begin{figure}[t]
  \centering
    \includegraphics[height=0.375\columnwidth]{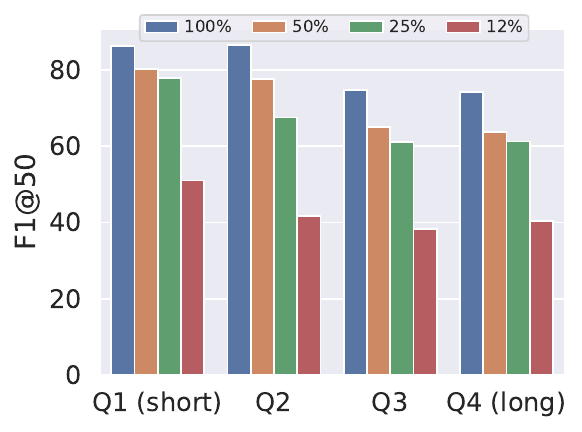}
    \includegraphics[height=0.375\columnwidth]{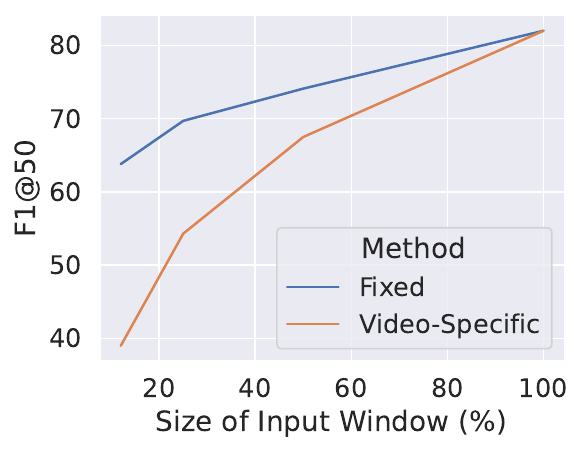}
    
  \vspace{-4mm}
  \caption{Results for different video lengths (rights). Comparison of a fixed with a video-specific input window size (left). Both plots are for 50Salads. 
  }
  \label{fig:video_spec_window_expr}
  \vspace{-4mm}
\end{figure}

We furthermore evaluated whether the impact of the window size is stronger for longer videos. To this end, we sorted all test videos into four quarters by their length. Fig.~\ref{fig:video_spec_window_expr} (left) shows that the difference between the window size 50\% and the full video (100\%) is larger for long videos. This shows that long-term context is in particular for long videos important. We also evaluated whether choosing a window for each video instead of choosing a window based on the average video length (fixed) performs better. For the video-specific window size, the window size is set to the percentage of each video. Fig.~\ref{fig:video_spec_window_expr} (right) shows that a video-specific window size performs much worse than a fixed window. Varying the amount of context for each video is thus not beneficial.      Fig.~\ref{fig:teaser_assembly} shows qualitative results of our approach for different amounts of temporal context with a fixed window for a video from the Assembly101 dataset.       

\begin{table}[t]
\centering
\resizebox{\linewidth}{!}{%
\setlength{\tabcolsep}{4pt}
\begin{tabular}{l | c | ccc|c|c || ccc|c|c }
\hline
 \multirow{2}{*}{Method} &  \multirow{2}{*}{Note} & \multicolumn{5}{c ||}{Breakfast} & \multicolumn{5}{c}{50Salads} \\
\cline{3-12}
  & & \multicolumn{3}{c|}{F1@\{10, 25, 50\}} & Edit & Acc 
   & \multicolumn{3}{c|}{F1@\{10, 25, 50\}} & Edit & Acc  \\
\hline
IDT+LM~\cite{richard2016temporal}
& DF\textsubscript{1}  
& - & - & - & - & -  
& 44.4 & 38.9 & 27.8 & 45.8 & 48.7  
\\
ST-CNN~\cite{lea2016segmental}
& DF\textsubscript{2}
& - & - & - & - & -  
& 55.9 & 49.6 & 37.1 & 45.9 & 59.4  
\\
ED-TCN~\cite{lea2017temporal}
& DF\textsubscript{2}
& - & - & - & - & -  
& 68.0 & 63.9 & 52.6 & 59.8 & 64.7  
\\
TDRN~\cite{lei2018temporal}
& DF\textsubscript{2}
& - & - & - & - & -  
& 72.9 & 68.5 & 57.2 & 66.0 & 68.1  
\\
SSA-GAN~\cite{gammulle2020fine}
& DF\textsubscript{3}
& - & - & - & - & 43.3  
& 74.9 & 71.7 & 67.0 & 69.8 & 73.3  
\\
Bridge-Prompt~\cite{li2022bridge}
& DF\textsubscript{4}
& - & - & - & - & - 
& 89.2 & \textbf{\textit{87.8}} & 81.3 &  \textbf{\textit{83.8}}  & \textbf{\textit{88.1}}
\\
\midrule
C2F-TCN~\cite{singhania2021} 
& TA
& 72.2 & 68.7 & 57.6 & 69.6 & \textbf{\textit{76.0}}  
& 84.3 & 81.8 & 72.6 & 76.4 & 84.9  
\\
UVAST~\cite{behrmann2022uvast}+Viterbi
& P
& 75.9  & 70.0 & 57.2 & 76.5 & 66.0 
& 89.1  & 87.6 & 81.7 & \textbf{\textit{83.9}} & 87.4
\\
UVAST~\cite{behrmann2022uvast}+FIFA~\cite{FIFA} 
& P
& 76.9  & 71.5 & 58.0 & 77.1 & 69.7 
& 88.9  & 87.0 & 78.5 & 83.9 & 84.5
\\
Liu \cite{liu2023temporal} + ASRF~\cite{ishikawa2021asrf}
& P
& 77.5 & 72.3 & 59.5 & 76.7 & 73.7 
& 87.9 & 86.6 & 80.5 &  82.7  & 86.6
\\ 
DTL~\cite{xudon2022dtl} 
& C 
& \textbf{\textit{78.8}} & \textbf{\textit{74.5}} & \textbf{\textit{62.9}} & \textbf{\textit{77.7}} & \textbf{\textit{75.8}} 
& 87.1 & 85.7 & 78.5 &  80.5  & 86.9
\\ 
\midrule\midrule
MS-TCN~\cite{farha2019ms}
& -
& 52.6 & 48.1 & 37.9 & 61.7 & 66.3  
& 76.3 & 74.0 & 64.5 & 67.9 & 80.7 
\\
MS-TCN++~\cite{li2020ms}
& -
& 64.1 & 58.6 & 45.9 & 65.6 & 67.6  
& 80.7 & 78.5 & 70.1 & 74.3 & 83.7  
\\
DTGRM~\cite{wang2020temporal}
& -
& 68.7 & 61.9 & 46.6 & 68.9 & 68.3  
& 79.1 & 75.9 & 66.1 & 72.0 & 80.0  
\\
MuCon~\cite{souri2021MuCon}
& -
& 73.2 & 66.1 & 48.4 & 76.3 & 62.8  
& - & - & - & - & -  
\\
Gao~\etal~\cite{gao2021global2local}
& -
& 74.9 & 69.0 & 55.2 & 73.3 & 70.7  
& 80.3 & 78.0 & 69.8 & 73.4 & 82.2  
\\
BCN~\cite{wang2020boundary}
& -
& 68.7 & 65.5 & 55.0 & 66.2 & 70.4  
& 82.3 & 81.3 & 74.0 & 74.3 & 84.4  
\\
SSTDA~\cite{chen2020action}
& -
& 75.0 & 69.1 & 55.2 & 73.7 & 70.2  
& 83.0 & 81.5 & 73.8 & 75.8 & 82.2  
\\
C2F-TCN~\cite{singhania2021} 
& -
& 70.1 & 66.6 & 56.2 & 68.2 & 73.5  
& 76.6 & 73.0 & 62.5 & 69.2 & 80.1  
\\
ASRF~\cite{ishikawa2021asrf} 
& -
& 74.3 & 68.9 & 56.1 & 72.4 & 67.6  
& 84.9 & 83.5 & 77.3 & 79.3 & 84.5 
\\
UVAST~\cite{behrmann2022uvast} 
& -
& \underline{76.7} & 70.0 & 56.6 & \textbf{77.2} & 68.2 
& 86.2  & 81.2 & 70.4 & \textbf{83.9} & 79.5
\\
DPRN~\cite{park2022maximization} 
& -
& 75.6 & 70.5  & 57.6 & 75.1 & 71.7 
& \underline{87.8} & \underline{86.3}  & 79.4 & 82.0 & \underline{87.2} 
\\
LGTNN~\cite{tian2022lgtnn} 
& -
& 76.2 & \underline{71.5} & 57.5 & 75.2 & 72.5 
& 87.5 & 86.2  & 79.8 & 82.0 & 86.1
\\
ASFormer~\cite{yi2021asformer}
& -
& 76.0 & 70.6  & 57.4 & 75.0 & 73.5 
& 85.1 & {83.4}  & 76.0 & 79.6 & 85.6
\\
TCTr~\cite{aziere2022multistage} 
& -
& 76.6 & 71.1 & \underline{58.5} & 76.1 & \textbf{77.5}  
& 87.5 & 86.1 & \underline{80.2} & \underline{83.4} & 86.6 
\\
\midrule
LTContext (Ours)
& -
& \textbf{77.6} & \textbf{72.6} & \textbf{60.1} &  \underline{77.0}  & \underline{74.2}  
& \textbf{89.4} & \textbf{87.7} & \textbf{82.0} & 83.2 & \textbf{87.7} 
\\
\bottomrule
\end{tabular}
}
\caption{
Results on the Breakfast and 50Salads datasets. The best and second best results for the methods in the bottom half are shown in bold and underlined since only methods without additional notes are directly comparable. P: additional post-processing; C: additional constraints; DF: different features (DF\textsubscript{1}: Improved Dense Trajectories (IDT), DF\textsubscript{2}: Spatio-temporal VGG-style CNN, DF\textsubscript{3}: Generative Adversarial Network (GAN), DF\textsubscript{4}: ActionCLIP); TA: test augmentation. In the top half, we highlight results that are better in italic bold formatting.
}
\vspace{-0.05in}
\label{tab:results_breakfast_salasds}
\end{table}

\begin{table}[t]
\centering
\resizebox{\columnwidth}{!}{%
\begin{tabular}{l | ccc|c|c | c | c }
\hline
 \multirow{2}{*}{Method}  & \multicolumn{7}{c}{Assembly101} \\
\cline{2-8}
  &  \multicolumn{3}{c|}{F1@\{10, 25, 50\}} & Edit & Acc & Params (M) & Inference (sec) \\
\hline
MS-TCN++~\cite{li2020ms}          &  31.6 & 27.8 & 20.6 &  30.7 & 37.1 &  1.08 & 0.1  \\
UVAST~\cite{behrmann2022uvast}*   & 32.1  & 28.3 & 20.8 &  \underline{31.5} & 37.4 &  1.22 & 3.2 \\
C2F-TCN~\cite{singhania2021}      & 33.3  & 29.0 & 21.3 & \textbf{32.4} & \underline{39.2} & 6.89 & 0.1 \\
ASFormer~\cite{yi2021asformer}*                         &  \underline{33.4}   & \underline{29.2} & \underline{21.4} &  30.5 & 38.8           &  1.13 & 2.1 \\
\midrule
LTContext (Ours)                              & \textbf{33.9} & \textbf{30.0} & \textbf{22.6} & 30.4 & \textbf{41.2} & 0.72 & 0.6 \\
\bottomrule
\end{tabular}
}
\caption{
Comparison with state-of-the-art methods on the Assembly101 dataset. The best and second best results are shown in bold and underlined. *We trained UVAST~\cite{behrmann2022uvast} and ASFormer~\cite{yi2021asformer} using the code of the authors.
}
\vspace{-0.05in}
\label{tab:results_assembly101}
\end{table}

\subsection{Comparison with State of the Art}
We present the performance comparison of our method with state-of-the-art methods on the Breakfast and 50Salads datasets in Table~\ref{tab:results_breakfast_salasds} and on the Assembly101 dataset~\cite{sener2022assembly101} in Table~\ref{tab:results_assembly101}.

 On Breakfast and 50Salads, our method outperforms all comparable methods in terms of F1 score at all thresholds, which is the most important measure. Our method also achieves a better Edit score than all methods except for UVAST~\cite{behrmann2022uvast} and TCTr~\cite{aziere2022multistage}. UVAST uses ASFormer~\cite{yi2021asformer} as encoder and an additional alignment decoder. The higher Edit score of UVAST is expected since the approach has an additional head that predicts the sequence of actions and thus maximizes the Edit score by an additional loss. This, however, comes at the cost of a much lower frame accuracy. TCTr~\cite{aziere2022multistage} achieves a higher frame-wise accuracy on Breakfast and a higher Edit score on 50 Salads, but it performs worse for the other metrics. The approach uses a boundary detection module and compresses the temporal features, which is a complementary approach. It needs to be noted that our approach outperforms TCTr for all metrics on 50Salads if we use 10 instead of 9 layers as shown in Table~\ref{tab:ablation_num_layers}. However, 9 layers perform better than 10 layers for the other datasets. We therefore report in Tables~\ref{tab:results_breakfast_salasds} and \ref{tab:results_assembly101} only the results with 9 layers.     

 We also compare to approaches that use different input features, perform additional test augmentation, or add additional constraints. On Breakfast, only DTL~\cite{xudon2022dtl} reports better results. DTL~\cite{xudon2022dtl} uses additional logic-based constraints for training ASFormer and assumes that action sequences can be modeled by a directed acyclic graph. This is very efficient on the Breakfast dataset, which has relatively few actions per video compared to the other datasets. However, our approach outperforms DTL for all metrics on 50Salads where the structure of the action sequences is more complex. Furthermore, DTL cannot be applied to datasets like Assembly101 where actions occur more than once in a video. 
 On 50Salads, only Bridge-Prompt~\cite{li2022bridge} performs slightly better for some metrics. Bridge-Prompt proposes an approach for feature learning using a vision-language model such as ActionCLIP~\cite{wang2021actionclip} in combination with ASFormer. The approach is thus complementary and not comparable to our approach. Nevertheless, our approach achieves higher $F1@10$ and $F1@50$ scores on 50Salads. If we use 10 instead of 9 layers as shown in Table~\ref{tab:ablation_num_layers}, we even outperform Bridge-Prompt for all measures except for frame-wise accuracy. In summary, our approach outperforms all methods in terms of F1 score, which is the most precise measure of the segmentation quality, at all thresholds on 50Salads.

 Assembly101 is the largest dataset both in terms of the number of videos and their length, and it is the most challenging dataset. Since ASFormer~\cite{yi2021asformer} and UVAST~\cite{behrmann2022uvast} have not been evaluated on this dataset, we trained ASFormer and UVAST on Assembly101 using the publicly available source code and report the results in Table~\ref{tab:results_assembly101} as well. We outperform all methods on Assembly101 in F1 scores. C2F-TCN~\cite{singhania2021} achieves the best edit score.
 Since we trained ASFormer~\cite{yi2021asformer} and UVAST~\cite{behrmann2022uvast} by ourselves, we can compare the training time on Assembly101. While our approach requires 1 day and 18 hours, ASFormer~\cite{yi2021asformer} and UVAST~\cite{behrmann2022uvast} needed 4 weeks and 2 weeks, respectively.

\begin{figure*}[t]
  \centering
  \setlength{\abovecaptionskip}{0pt plus 1pt minus 1pt}
\begin{subfigure}[b]{1.0\linewidth}
    \centering
    \includegraphics[width=1.0\linewidth]{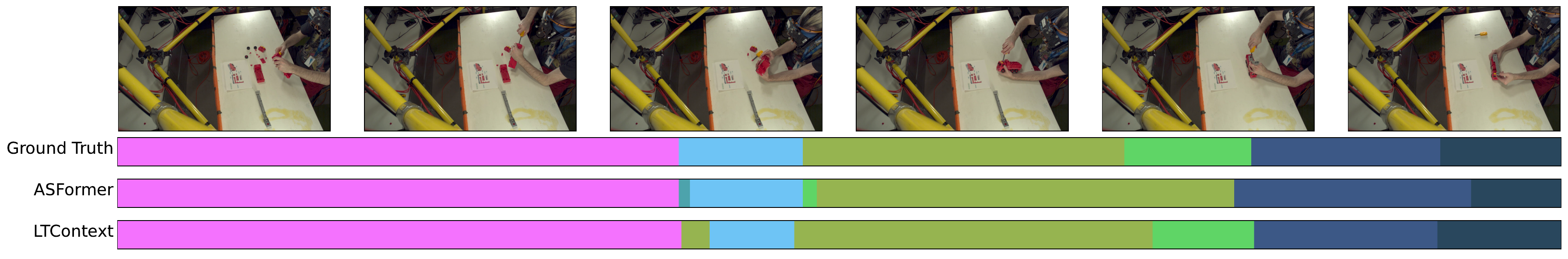}
    \label{fig:figure1}
  \end{subfigure}\hspace{-10mm}
  \begin{subfigure}[b]{1.0\linewidth}
    \centering
    \includegraphics[width=1.0\linewidth]{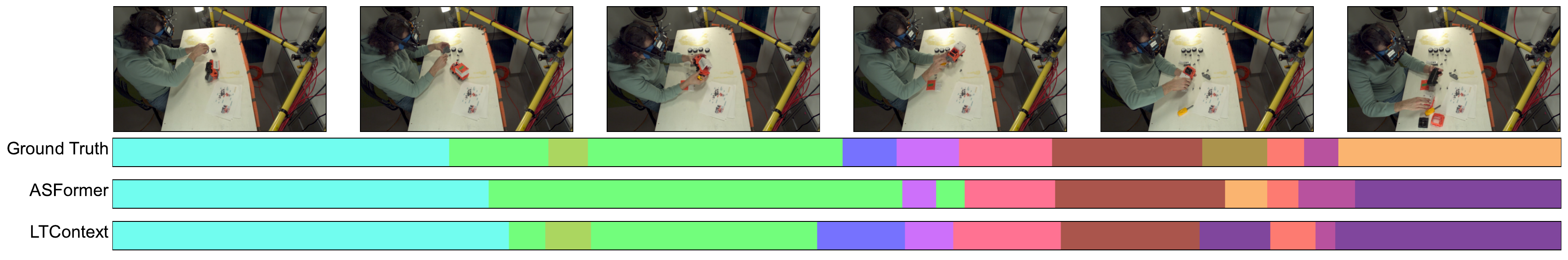}
    \label{fig:figure2}
  \end{subfigure}
  \caption{Qualitative results on Assembly101. The three rows show the ground-truth labels, the predictions by ASFormer, and the predictions by the proposed approach LTContext. It can be best viewed by using the zoom function of a PDF viewer.
  }
  \label{fig:qual_assembly}
\end{figure*}

\begin{figure*}[t]
    \centering
    \includegraphics[width=1.0\linewidth]{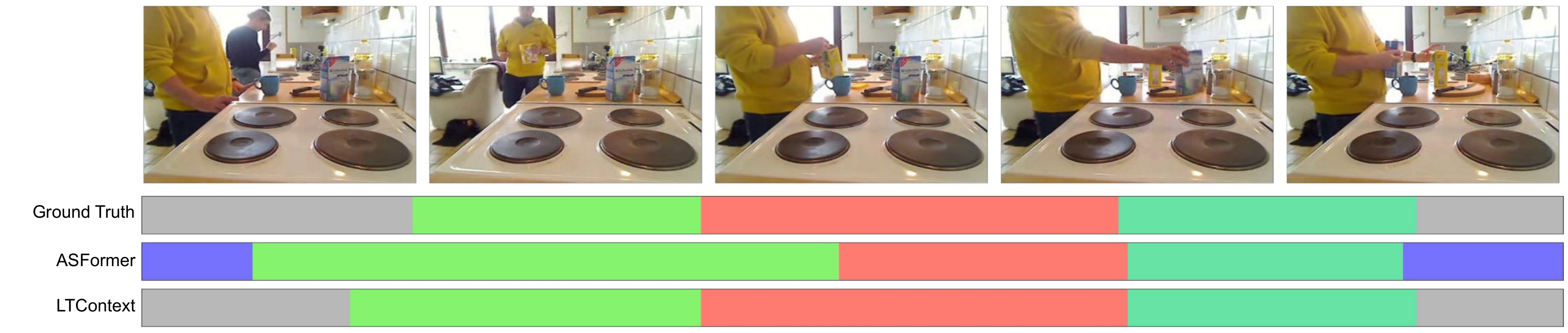}
    \caption{Qualitative results on Breakfast. The three rows show the ground-truth labels, the predictions by ASFormer, and the predictions by the proposed approach LTContext.}
    \label{fig:qual_breakfast}
\end{figure*}

\begin{figure*}[t]
    \centering
    \includegraphics[width=.9\linewidth]{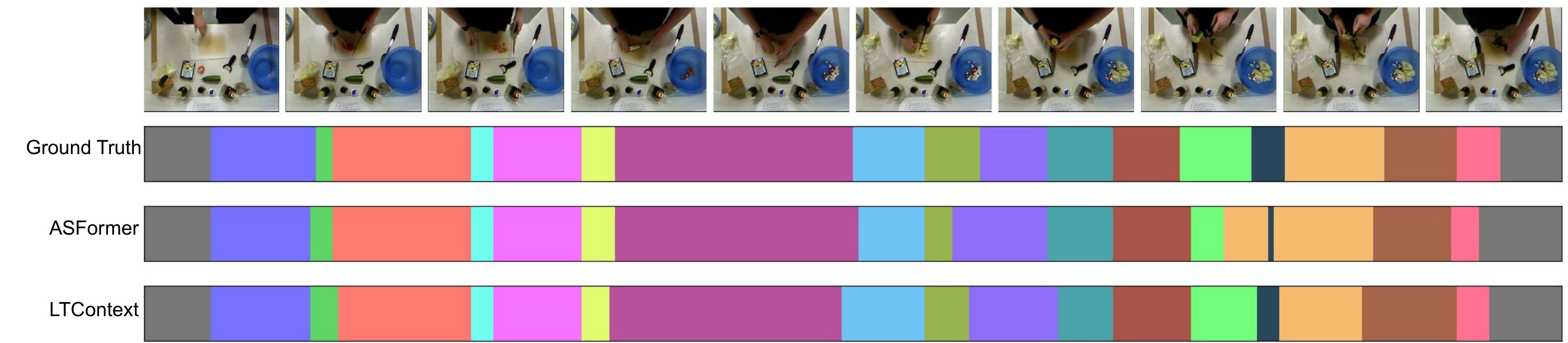}
    \caption{Qualitative results on 50Salads. The three rows show the ground-truth labels, the predictions by ASFormer, and the predictions by the proposed approach LTContext.}
    \label{fig:qual_50salads}
\end{figure*}

\subsection{Qualitative Evaluation}
In Figs.~\ref{fig:qual_assembly}-\ref{fig:qual_50salads}, we present some qualitative results for the Assembly101, Breakfast, and 50Salads dataset. The first row of each figure shows the middle frame of each ground-truth action segment. In the second, third, and fourth rows, the ground truth segmentation, the predictions of ASFormer~\cite{yi2021asformer}, and the prediction of our model (LTContext) are shown, respectively. 
In Fig.~\ref{fig:qual_assembly}, our approach shows much fewer errors than ASFormer. Although our approach recognizes all action classes that occur in the video, there are several errors where some instances are missed like the last action segment of the bottom row which corresponds to the action \textit{`detach base'}. This indicates how challenging the Assembly101 dataset is. 
In Fig.~\ref{fig:qual_breakfast}, the predictions of our model are very close to the ground-truth. ASFormer overestimates the duration of the green segment, which corresponds to the action \textit{`spoon powder'}, and hallucinates purple segments at the beginning and end of the video. In Fig.~\ref{fig:qual_50salads}, both methods estimate the segments well, but ASFormer predicts wrongly two orange segments, which correspond to the action \textit{`mix dressing'}, and the olive segment is too short, which corresponds to the action \textit{`add salt'}.

\subsection{Ablation Studies}
We finally evaluate the impact of each component of our architecture. For the ablation studies, we report the results averaged over the 5 splits of the 50Salads dataset.

\begin{table}[t]
\centering
\resizebox{\columnwidth}{!}{%
\begin{tabular}{l | ccc cc }
\toprule
Model Architecture & \multicolumn{3}{c}{F1@\{10, 25, 50\}} & Edit & Acc \\
\midrule
LTContext & \textbf{89.4} & \textbf{87.7} & \textbf{82.0} & \textbf{83.2} & \textbf{87.7} \\
\hspace{2px} - Windowed Attention  & 87.7 & 85.1 & 77.4 & 82.1 & 85.5  \\
\hspace{2px} - LTContext Attention & 84.1 & 82.6 & 74.1 & 78.2 & 85.3  \\
WA (S1) + LTContext (S2-S4)  & 87.8 & 85.0 & 79.4 & 81.4 & 85.2  \\
\bottomrule
\end{tabular}
}
\caption{
Impact of using LTContext and Windowed Attention (WA) on 50Salads. In the case of Windowed Attention, we use two windowed attention blocks instead of a combination of windowed and LTContext attention. In the case of LTContext Attention, we use two LTContext attention blocks.
The last row corresponds to using only windowed attention in stage 1 and only LTContext attention in stages 2 to 4.
}
\vspace{-0.05in}
\label{tab:ablation_attentions}
\end{table}

\paragraph{Impact of attention types} 
In Table~\ref{tab:ablation_attentions}, we show the impact of using the combination of windowed and long-term context (LTContext) attention in the LTContext block illustrated in Fig.~\ref{fig:ltc_arch}. We first compare it to two variants where we use only windowed or only long-term context attention. In order to keep the number of parameters the same, we still use in these cases two attention blocks within LTContext. The results show that combining both types of attention leads to better results. In particular, the F1@50 score is substantially higher.  

As shown in Fig.~\ref{fig:ltc_arch}, we use four LTContext blocks. We also evaluate what happens if we vary the attention not within an LTContext block but between the four LTContext blocks. For this, we use only windowed attention for the first LTContext block and only long-term context attention for the other three LTContext blocks. The last row in Table~\ref{tab:ablation_attentions} shows that this does not perform better than using only windowed attention and it is worse than combining windowed and long-term context attention within an LTContext block.

\paragraph{Impact of different values of $W$ and $G$}
The parameter $W$ controls the size of the local window for the local attention and the parameter $G$ controls the sparseness of the global attention. If not otherwise specified, we use $W=G=64$ in our experiments. The results in Table~\ref{tab:ablation_window_size} show that the F1 score drops when the size of the local window $W$ becomes smaller. Note that larger values of $W$ increase the memory and computational cost. When we decrease $G$, the F1 score also drops but not as drastic as for $W$.

\begin{table}[t]
\centering
\resizebox{0.8\columnwidth}{!}{%
\begin{tabular}{l l | ccc cc }
\toprule
$W$ & $G$ & \multicolumn{3}{c}{F1@\{10, 25, 50\}} & Edit & Acc \\
\midrule
8  &  64  & 83.8 & 82.1 & 74.5 & 77.0 & 87.0  \\
16 &  64  & 85.4 & 82.9 & 77.1 & 77.1 & 87.0  \\
32 &  64  & 87.6 & 85.8 & 79.8 & 81.5 & 86.6  \\
64 &  64 & \textbf{89.4} & \textbf{87.7} & \textbf{82.0} & \textbf{83.2} & \textbf{87.7} \\
64 &  32 & 88.8 & 87.2 & 81.3 & 83.2 & 87.7  \\
64 &  16 & 88.6 & 87.0 & 80.3 & 82.6 & 86.2  \\
64 &  8  & 88.6 & 87.1 & 79.9 & 82.9 & 86.6  \\
\bottomrule
\end{tabular}
}
\caption{
Impact of the parameters $W$ and $G$ on 50 Salads.
}
\vspace{-0.05in}
\label{tab:ablation_window_size}
\end{table}

\paragraph{Impact of overlaps for windowed attention}
As described in Section 3.1, we use an overlap for the keys and values for the windowed attention. In Table~\ref{tab:ablation_block_overlap}, we also report the result when the keys and values do not overlap, \ie, they are the same as the queries. It is interesting to note that our attention can be interpreted as a combination of a reshaping operation with axial attention~\cite{ho2019axial} in the special case without overlap. The results, however, show that an overlap improves the results. Using a larger overlap where the keys and values consist of three consecutive windows does not improve the results further.

\begin{table}[t]
\centering
\resizebox{0.85\columnwidth}{!}{%
\begin{tabular}{c | ccc cc }
\toprule
 & \multicolumn{3}{c}{F1@\{10, 25, 50\}} & Edit & Acc \\
\midrule
Non-Overlapping   & 87.9 & 85.5 & 79.1 & 81.4 & 85.1  \\
  Overlap (2 Windows)       & \textbf{89.4} & 87.7 & \textbf{82.0} & \textbf{83.2} & \textbf{87.7}  \\
    Overlap (3 Windows)       & 89.2 & \textbf{88.0} & 80.8 & \textbf{83.2} & 86.6  \\
\bottomrule
\end{tabular}
}
\caption{
Impact of using an overlap for the keys and values for the local attention on 50Salads.
}
\vspace{-0.05in}
\label{tab:ablation_block_overlap}
\end{table}

\paragraph{Impact of the order of the attention}
The LTContext block shown in Fig.~\ref{fig:ltc_arch} uses first windowed attention and then long-term context attention.
Table~\ref{tab:ablation_attn_order} shows the results when we change the order of windowed and long-term context attention within the LTContext block. If the order is changed, the performance decreases. Since the LTContext blocks are repeated, the drop in performance is moderate. 

\begin{table}[t]
\centering
\resizebox{0.9\columnwidth}{!}{%
\begin{tabular}{c | ccc cc }
\toprule
 & \multicolumn{3}{c}{F1@\{10, 25, 50\}} & Edit & Acc \\
\midrule
  WA-LTContext       & \textbf{89.4} & \textbf{87.7} & \textbf{82.0} & \textbf{83.2} & \textbf{87.7}  \\
  LTContext-WA       & 88.3 & 86.4 & 80.4 & 81.6 & 86.1  \\
\bottomrule
\end{tabular}
}
\caption{
Impact of the attention order within the LTContext block on 50Salads. Our model computes first windowed attention (WA) and then long-term context attention (row 1).
}
\vspace{-0.05in}
\label{tab:ablation_attn_order}
\end{table}

\paragraph{Impact of the cross-attention}
As described in Section \ref{sec:deglas_arch}, we use cross-attention in stages 2 to 4. Table \ref{tab:ablation_cross_attn} shows that the performance drastically decreases without cross-attention.
\begin{table}[t]
\centering
\resizebox{0.9\columnwidth}{!}{%
\begin{tabular}{c | ccc cc }
\toprule
 & \multicolumn{3}{c}{F1@\{10, 25, 50\}} & Edit & Acc \\
\midrule
  Without Cross-Attention       & 82.2 & 78.5 & 70.4 & 73.6 & 80.2  \\
LTContext       & \textbf{89.4} & \textbf{87.7} & \textbf{82.0} & \textbf{83.2} & \textbf{87.7}  \\
\bottomrule
\end{tabular}
}
\caption{
Impact of the cross-attention on 50Salads.
}
\vspace{-0.05in}
\label{tab:ablation_cross_attn}
\end{table}

\paragraph{Impact of using convolutions}
The LTContext block shown in Fig.~\ref{fig:ltc_arch} starts with a dilated 1D convolution with kernel size 3. In Table~\ref{tab:ablation_conv}, we evaluate the impact of the dilated convolution by comparing it to a LTContext block that uses a convolutional kernel of the same size but without dilation factor, and a LTContext block without 1D convolution. The results show that dilated convolutions have a very high impact on performance.

\begin{table}[t]
\centering
\resizebox{.9\columnwidth}{!}{%
\begin{tabular}{c | ccc cc }
\toprule
 & \multicolumn{3}{c}{F1@\{10, 25, 50\}} & Edit & Acc \\
\midrule
  Conv with Dilation       & \textbf{89.4} & \textbf{87.7} & \textbf{82.0} & \textbf{83.2} & \textbf{87.7}  \\
  Conv without Dilation    & 80.1 & 78.0 & 67.7 & 72.6 & 79.3  \\
  Without Conv             & 81.8 & 79.2 & 67.1 & 72.7  & 78.9  \\
\bottomrule
\end{tabular}
}
\caption{
Impact of using LTContext blocks with 1D convolution but without dilation (row 2) and using LTContext blocks without 1D convolution (row 3).
}
\vspace{-0.05in}
\label{tab:ablation_conv}
\end{table}

\paragraph{Impact of the number of layers} 
As shown in Fig.~\ref{fig:ltc_arch}, we repeat the LTContext blocks at each stage $N$ times. We used $N{=}9$ layers in all experiments. Table~\ref{tab:ablation_num_layers} shows the impact of varying the number of layers. On 50Salads, all measures improve by increasing $N$. It needs to be noted that we use $N{=}9$ for the results reported in Table~\ref{tab:results_breakfast_salasds} although we can get even better results with 10 layers on 50 Salads. For the Breakfast and Assembly101 dataset, the best performance is achieved with 9 layers.

\begin{table}[t]
\centering
\resizebox{.9\columnwidth}{!}{%
\begin{tabular}{c | ccc cc  }
\toprule
 \multirow{2}{*}{number of layers (N)} & \multicolumn{5}{c}{50Salads}  \\
  & \multicolumn{3}{c}{F1@\{10, 25, 50\}} & Edit & Acc \\
\midrule
  8   & 87.2 & 85.1 & 77.6 & 80.8 & {85.7}  \\
  9   & {89.4} & {87.7} & {82.0} & {83.2} & \textbf{87.7}  \\
  10  & \textbf{89.5} & \textbf{88.1} & \textbf{82.4} & \textbf{84.1} &  \textbf{87.7}  \\
  \midrule
 &  \multicolumn{5}{c}{Breakfast} \\
\midrule
8 & 75.7 & 70.5 & 57.7 & 74.5 & 73.1  \\
9 & \textbf{77.6} & \textbf{72.6} & \textbf{60.1} & \textbf{77.0} & \textbf{74.2}  \\
10 & 77.3 & 72.4 & 59.7 & 76.4 & 73.5 \\ 
\midrule
 &  \multicolumn{5}{c}{Assembly101} \\
 \midrule
8 & 31.9 & 28.4 & 21.3 & 27.8 & 41.0  \\
9 & \textbf{33.9} & \textbf{30.0} & \textbf{22.6} & \textbf{30.4} & 41.2  \\
10 & 32.6 & 29.3 & 21.9 & 28.7 & \textbf{41.5}  \\
\bottomrule
\end{tabular}
}
\caption{
Impact of the number of layers on 50Salads, Breakfast, and Assembly101. 
}
\vspace{-0.05in}
\label{tab:ablation_num_layers}
\end{table}

\paragraph{Impact of the number of attention heads} In our implementation, we do not use multiple attention heads. Nevertheless, we evaluated the impact of using multiple heads in Table~\ref{tab:ablation_num_attn_heads} since most transformers use multiple heads. The results, however, show that there is no benefit in using multiple heads. The results are consistent with the observations in~\cite{michel2019heads, liu2021heads}. For example, \cite{michel2019heads} shows that many attention heads can often be reduced to a single head without impacting the performance. They also argue that some tasks are more reliant on multiple heads than others. Temporal action segmentation seems to be a task where one head is sufficient.   

\begin{table}[t]
\centering
\resizebox{.9\columnwidth}{!}{%
\begin{tabular}{c | ccc cc }
\toprule
number of heads & \multicolumn{3}{c}{F1@\{10, 25, 50\}} & Edit & Acc \\
\midrule
  1   & \textbf{89.4} & \textbf{87.7} & \textbf{82.0} & \textbf{83.2} & \textbf{87.7}  \\
  2   & 88.8 & 87.0 & 80.5 & 82.9 & 87.0  \\
  4   & 89.2 & 87.1 & 80.8 & 83.1 & 86.9  \\
  8   & 88.2 & 86.4 & 80.5 & 82.1 & 86.1  \\
\bottomrule
\end{tabular}
}
\caption{
Impact of the number of attention heads.
}
\vspace{-0.05in}
\label{tab:ablation_num_attn_heads}
\end{table}

\paragraph{Impact of the number of stages} As shown in Fig.~\ref{fig:ltc_arch}, we use four stages of LTContext blocks, each of them with 9 layers. We evaluate the impact of the number of stages in Table~\ref{tab:ablation_num_stages}. As can be seen, using multiple stages helps to reduce the over-segmentation error and improves the F1 score and Edit score significantly compared to using only one stage. Increasing the number of stages up to 4 improves all metrics, but using 5 stages decreases the performance and the network starts to overfit. 

\begin{table}[t]
\centering
\resizebox{.85\columnwidth}{!}{%
\begin{tabular}{c | ccc cc }
\toprule
number of stages & \multicolumn{3}{c}{F1@\{10, 25, 50\}} & Edit & Acc \\
\midrule
  1   & 56.3 & 54.1 & 49.9 & 45.6 & 84.7  \\
  2   & 86.2 & 84.5 & 78.2 & 80.5 & 86.7  \\
  3   & 87.6 & 85.7 & 78.3 & 81.2 & 85.5  \\
  4   & \textbf{89.4} & \textbf{87.7} & \textbf{82.0} & \textbf{83.2} & \textbf{87.7}  \\
  5   & 89.0 & 86.8 & 80.0 & 83.0 & 85.7  \\
\bottomrule
\end{tabular}
}
\caption{
Impact of the number of stages on 50Salads.
}
\vspace{-0.05in}
\label{tab:ablation_num_stages}
\end{table}

\section{Conclusion}
\label{sec:conclusion}
In this work, we addressed the question of how much temporal long-term context is needed for action segmentation. Our analysis indicates that allowing networks to operate on the full input sequence is more beneficial compared to the case where the model has only access to a subset of the input. Based on our analysis, we presented LTContext, an approach for temporal action segmentation, where we leverage sparse attention to capture the long-term context of a video and windowed attention to model the local information in the neighboring frames. Our approach achieves state-of-the-art segmental F1 scores on the 50Salads and Assembly101 datasets, which contain long videos. 

\section*{Acknowledgement}

Juergen Gall has been supported by the Deutsche Forschungsgemeinschaft (DFG, German Research Foundation) GA 1927/4-2 (FOR 2535 Anticipating Human Behavior), the project iBehave (receiving funding from the programme “Netzwerke 2021”, an initiative of the Ministry of Culture and Science of the State of Northrhine Westphalia), by the Federal Ministry of Education and Research (BMBF) under grant no. 01IS22094E WEST-AI, and the ERC Consolidator Grant FORHUE (101044724). The sole responsibility for the content of this publication lies with the authors. 

{\small
\bibliographystyle{ieee_fullname}
\bibliography{main}
}

\appendix
\label{sec:appendix}
\section*{Appendix}

We provide additional experiments and implementation details.

\section{Implementation Details}
As mentioned in the paper, we use 9 layers and 4 stages for all datasets. We use $W=G=64$ for Assembly101 and 50Salads, and $W=64$ and $G=8$ for the Breakfast dataset since the videos are shorter than Assembly101 and 50Salads. We use Adam~\cite{kingma2014adam} optimizer and cosine learning rate decay~\cite{loshchilov17decay}. The starting learning rate for Breakfast and Assembly101 is 0.00025 and the decay to 0.00005 starts after 15 epochs. We train Breakfast for 150 epochs and Assembly101 for 120 epochs. The model for 50Salads is trained for 200 epochs with a fixed learning rate of 0.00065. 

\section{Impact of Temporal Downsampling}
\begin{figure}[h]
    \centering
    \includegraphics[width=\columnwidth]{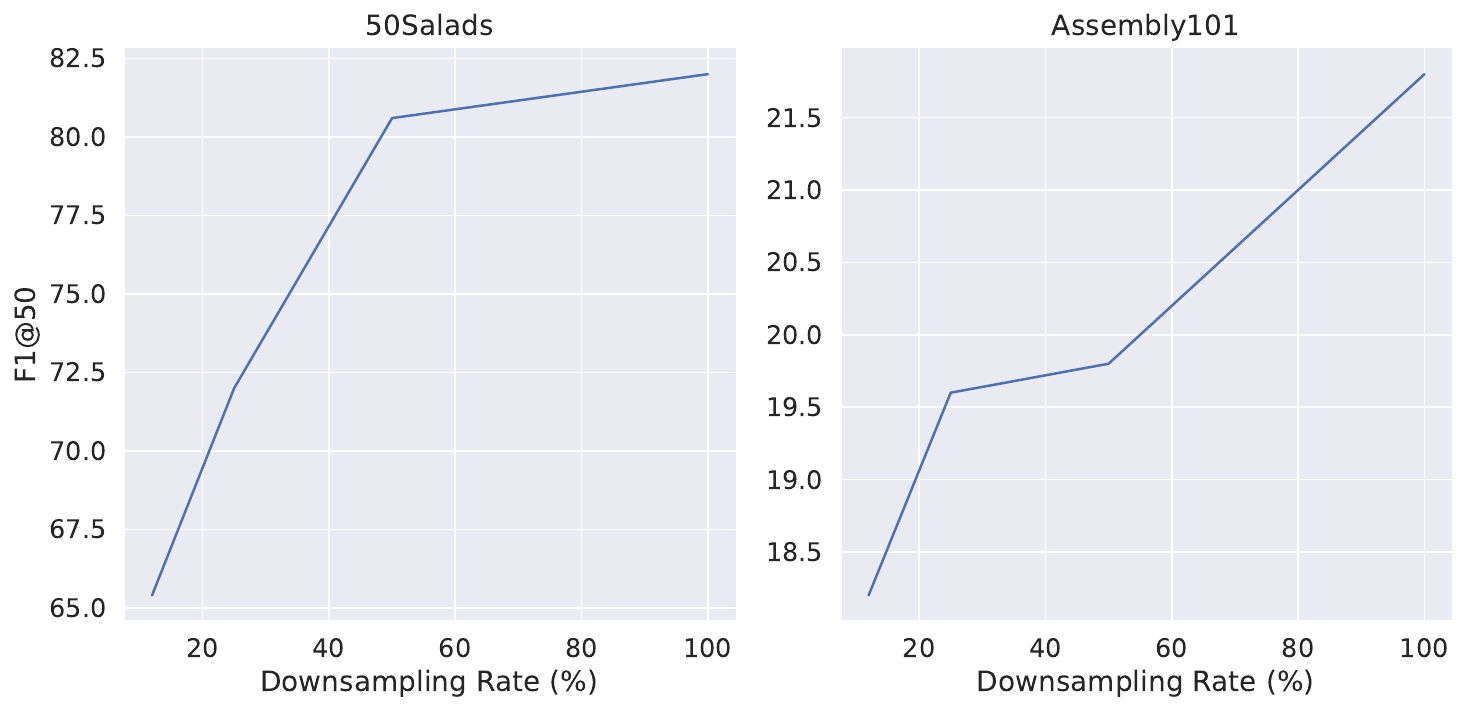}
    \caption{Impact of different downsampling rates on the 50Salads dataset (left) and the Assembly101 dataset (right).}
    \label{fig:downsample_expr}
\end{figure}

Fig.~\ref{fig:downsample_expr} shows the impact of temporally downsampling the input. In this experiment, the model has access to the full context of a video but in a lower temporal resolution since the input is temporally downsampled. The performance of the model degrades compared to no downsampling.

\section{Other Features}
In order to evaluate the impact of using vision-language models, we extract features using CLIP~\cite{radford21clip} from 50Salads and report the result of action segmentation in Table~\ref{tab:clip_feat}. Without additional fine-tuning, the features do not perform well.   

\begin{table}[h]
\centering
\resizebox{0.85\columnwidth}{!}{%
\begin{tabular}{c | ccc cc }
Features & \multicolumn{3}{c}{F1@\{10, 25, 50\}} & Edit & Acc \\
\hline
  CLIP        & 65.8 & 57.6 & 44.2 & 64.2 & 62.4  \\
  I3D        & 89.4 & 87.7 & 82.0 & 83.2 & 87.7  \\
  \hline
\end{tabular}
}
\vspace{-2mm}
\caption{
Results are on 50Salads.
}
\label{tab:clip_feat}
\end{table}

\section{Alternative Efficient Attentions}
We compare in Table \ref{tab:alternative_attn} our approach with RandomAttention~\cite{zaheer2020bigbird} from XFormer~\cite{xFormers2022} and FlashAttention~\cite{dao2022flashattention}. These types of attention focus on sparseness and result in fragmented segments, which is indicated by high accuracy, but very low F1
and Edit scores.
\begin{table}[h]
\centering
\resizebox{\columnwidth}{!}{%
\begin{tabular}{c | ccc cc }
 Attention & \multicolumn{3}{c}{F1@\{10, 25, 50\}} & Edit & Acc \\
\hline
  FlashAttention       & 55.2 & 53.0 & 48.7 & 42.6 & 84.6  \\
  RandomAttention       & 49.0 & 45.7 & 41.8 & 37.2 & 85.6  \\
  Ours       & 89.4 & 87.7 & 82.0 & 83.2 & 87.7  \\
  \hline
\end{tabular}
}
\vspace{-0mm}
\caption{
Results are on 50Salads.
}
\label{tab:alternative_attn}
\end{table}

\end{document}